\documentclass{article}
\usepackage{ijcai18}

\usepackage{times}
\usepackage{xcolor}
\usepackage{soul}
\usepackage[utf8]{inputenc}
\usepackage[small]{caption}
\usepackage{graphicx}
\usepackage{amssymb}
\usepackage{multirow}
\usepackage{amsmath}
\usepackage{epsfig}
\usepackage{amssymb}
\newcommand{\ancong}[1]{{\color{black}#1}}
\newcommand{\yinzhou}[1]{{\color{black}#1}}
\newcommand{\jason}[1]{{\color{black}#1}}
\newcommand{\CheckYinzhou}[1]{{\color{black}#1}}
\newcommand{\K}[1]{{\color{black}#1}}
\newcommand{\ja}[1]{{\color{black}#1}}
\newcommand{\yin}[1]{{\color{black}#1}}

\newcommand{\CameraReady}[1]{{\color{black}#1}}
\newcommand{\KK}[1]{{\color{black}#1}}

\author{
Zhou Yin$^1$,
Wei-Shi Zheng$^{1,3}$\thanks{Corresponding author, email: wszheng@ieee.org},
Ancong Wu$^1$, 
Hong-Xing Yu$^1$,
Hai Wan$^1$,\\
\textbf{Xiaowei Guo $^2$},
\textbf{Feiyue Huang $^2$},
\textbf{Jianhuang Lai $^1$}
\\
$^1$ School of Data and Computer Science, Sun Yat-sen University, Guangzhou, China\\
$^2$ YouTu Lab, Tencent\\
$^3$ Key Laboratory of Machine Intelligence and Advanced Computing, Ministry of Education, China\\
}
\title{Adversarial Attribute-Image Person Re-identification}

\begin{document}

\onecolumn{

\noindent \textbf{\LARGE{Adversarial Attribute-Image Person Re-identification}}

\vspace{2cm}

\noindent {\LARGE{Zhou Yin, Wei-Shi Zheng, Ancong Wu, Hong-Xing Yu, Hai Wan, Xiaowei Guo, Feiyue Huang, Jianhuang Lai}}

\Large
\vspace{2cm}

\noindent For reference of this work, please cite:

\vspace{1cm}
\noindent Adversarial Attribute-Image Person Re-identification
\\ Zhou Yin, Wei-Shi Zheng, Ancong Wu, Hong-Xing Yu, Hai Wan, Xiaowei Guo, Feiyue Huang, Jianhuang Lai, IJCAI, 2018
\vspace{1cm}

\noindent Bib:\\
\noindent
@article\{yin2018AIReID,\\
\ \ \   title=\{Adversarial Attribute-Image Person Re-identification\},\\
\ \ \  author=\{Zhou Yin, Wei-Shi Zheng, Ancong Wu, Hong-Xing Yu, Hai Wan, Xiaowei Guo, Feiyue Huang, Jianhuang Lai\},\\
\ \ \  journal=\{ International Joint Conference on Artificial Intelligence\},\\
\ \ \  year=\{2018\}\\
\}
\newpage

\maketitle

\begin{abstract}
\looseness=-1
While attributes have been widely used for person re-identification (Re-ID) \K{which} \KK{aims at matching} the same person images across disjoint camera views, they are used either as extra features or for performing multi-task learning to assist the image-image matching task. However, how to find a set of person images according to a given attribute description, which is very practical in many surveillance applications, remains a rarely investigated cross-modality matching problem in person Re-ID. In this work, we present this challenge and formulate this task as a joint space learning problem. 
By imposing an attribute-guided attention mechanism for images and a semantic consistent adversary strategy for attributes, each modality, i.e., images and attributes, successfully learns semantically correlated concepts under the guidance of the other. We conducted extensive experiments on three attribute datasets and demonstrated that the proposed joint space learning method is so far the most effective \K{method} for the attribute-image cross-modality person Re-ID problem.
\end{abstract}

%%%%%%%%% BODY TEXT
\section{Introduction}
\looseness=-1
Pedestrian attributes, e.g., age, gender and dressing, are searchable semantic elements to describe a person. 
\K{In many scenarios we need to search 
person images in surveillance environment according to specific attribute descriptions provided by users, as depicted in Figure \ref{OverallFigure}. We refer to this problem as} the \emph{attribute-image person re-identification \K{(attribute-image Re-ID)}}. This task is significant in finding missing people with tens of thousands of surveillance cameras equipped in modern metropolises. Compared with conventional image-based Re-ID \cite{reIDfeature,reIDmetric}, attribute-image Re-ID has the advantage that its query is much easier to be obtained, e.g., it is more practical to search for criminal suspects when only verbal testimony about the suspects' appearances is given.

%Attribute-based person re-identification  (Re-ID), which is depicted in Figure , aims at re-identifying 

\looseness=-1
Despite the great significance, the attribute-image Re-ID is still a very open problem and has been rarely investigated before. While a lot of attribute person Re-ID models \cite{lin2017improving,SVMAtt1,DBLP,AttributeReID1,AttributeReID2,AttributeReID3,AttributeReID4} have been developed recently, they are mainly used either for multi-task learning or for \K{providing} extra features so as to enhance the \CameraReady{performance of} image-image person Re-ID model. \K{The most intuitive solution to attribute-image Re-ID might be} to predict attributes for each person image, and search within the predicted attributes~\cite{RetrievalDependacy,AttributeSearch1,AttributeSpace}.
If we can reliably recognize the attributes of each pedestrian image, this might be the best way to find the person corresponding to the query attributes.
However, recognizing attributes from a person image is still an open issue, as pedestrian images from surveillance environment often suffer from low resolution, pose variations and illumination changes. The problem of imperfect recognition limits the intuitive methods \K{in bridging the gap between the two modalities (attribute and image), which are heterogeneous from each other. In addition,}
very often in a large-scale scenario, the predicted attributes from two pedestrians are different but very similar, \K{leading to} a very small inter-class distance in the predicted attribute space. Therefore, the imperfect prediction deteriorate the reliability of these existing models.

\begin{figure}
\setlength{\belowcaptionskip}{-0.5cm}
\begin{center}
\includegraphics[width=1.0\linewidth]{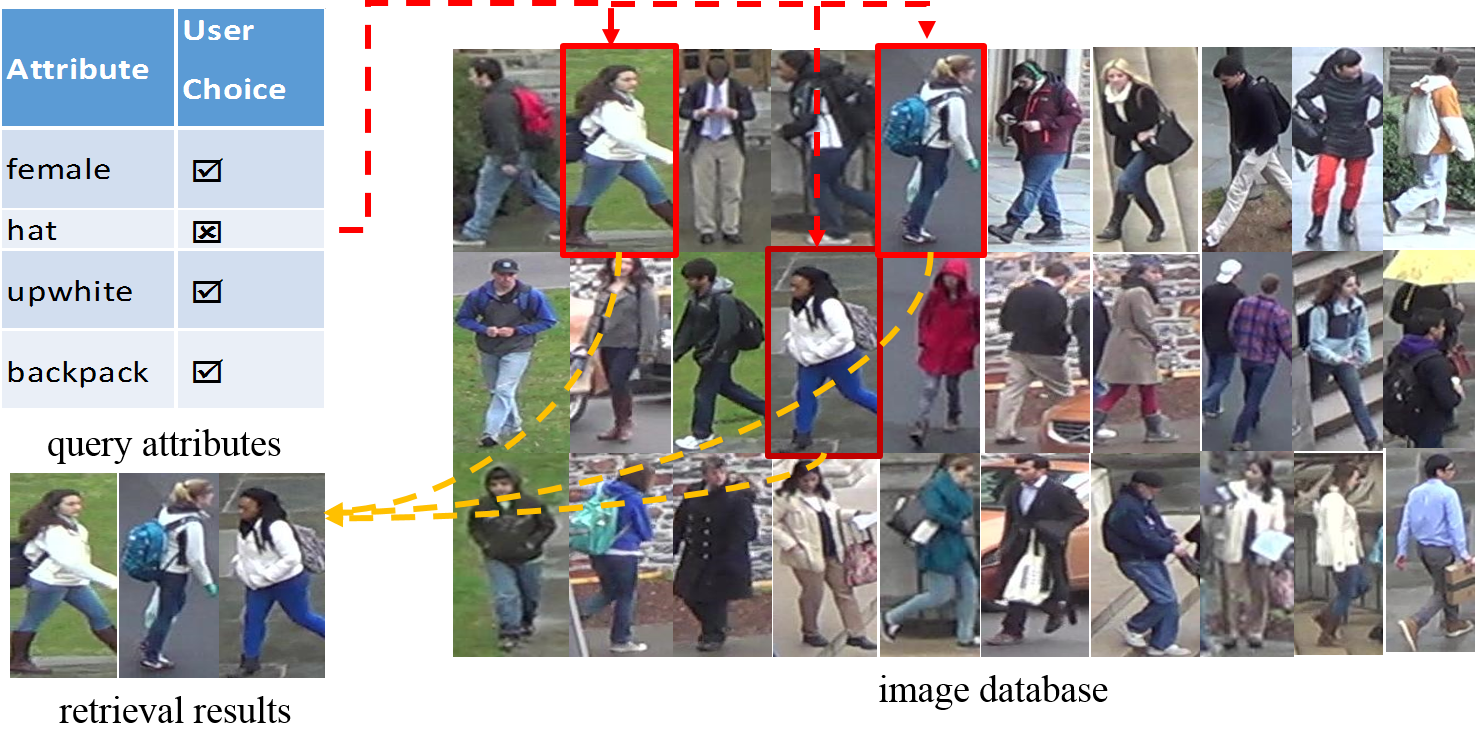}
\end{center}
\vspace{-0.5cm}
   \caption{The attribute-image Re-ID problem. The query is an attribute vector labeled by users, and the corresponding target person images that are matched with the query are retrieved.}
\label{OverallFigure}
\end{figure}
%very challenging due to the inconsistent distribution, as it requires to bridge the discriminability gap between the two intrinsically different modalities, i.e., between the high-level attribute which is abstract and structural and the low-level person image which is informative but noisy. 

%Such a problem is called the semantic gap problem.

\looseness=-1
In this paper, we \K{propose} an adversarial attribute-image Re-ID framework.
Intuitively, when we hold some attribute description in mind, e.g., ``dressed in red'', we generate an obscure and vague imagination on how a person dressed in red may look like, which we refer to as a concept. Once a concrete image is given, our vision system automatically processes the low-level features (e.g., color and edge) to obtain some perceptions, and then try to judge whether the perceptions and the concept are \K{consistent} with each other.

\looseness=-1
\K{More formally, \CameraReady{the goal of} our adversarial attribute-image Re-ID framework} is to learn a semantically discriminative joint space, which is regarded as a concept space (Figure \ref{network}), for generative adversarial architecture to generate some concepts \K{that are very similar to the concepts extracted from raw person images}.
\jason{However, the generic adversarial architecture is still hard to fit the match between two extremely large discrepant modalities (attribute and image). For this problem, we impose a semantic consistency regularization across modalities in order to regularize the adversarial architecture, \K{enhancing the learned joint space to build a bridge between the two modalities}.}

\looseness=-1
In a word, our framework learns \yin{a} semantically discriminative structure of low-level person images,
and generate \yin{a} correspondingly aligned image-analogous concept for high-level attribute towards image concept.
By the proposed strategy, directly estimating the attributes of a person image is averted, and the problems of imperfect prediction and low semantic discriminability are naturally solved,
because we learn a semantically discriminative joint space, rather than predicting and matching attributes.

We have conducted experiments on three large-scale benchmark datasets, namely Duke Attribute \cite{lin2017improving}, Market Attribute \cite{lin2017improving} and PETA \cite{PETA},
to validate our model. By our study, some interesting findings are:\\
%\ancong{/*** ancong: Summarize the contributions in experiments? It seems that it is proposed by trying in experiments.***/}
%\begin{itemize}
(1) Compared with other related cross-modality models, we find the regularized adversarial learning framework is so far most effective for solving the attribute-image Re-ID problem.\\
(2) For achieving better cross-modality matching between attribute and person image, it is more effective to use adversarial model to generate image-analogous concept and get it matched with image concept rather than doing \K{this in reverse}.\\
(3) The semantic consistency as regularization on  adversarial learning is important for the attribute-image Re-ID.
%\end{itemize}

%-------------------------------------------------------------------------
\section{Related Works}

\subsection{Attribute-based Re-ID}
\looseness=-1
While pedestrian attributes \K{in most research are} side information or mid-level representation to improve conventional image-image Re-ID tasks \cite{lin2017improving,SVMAtt1,DBLP,AttributeReID1,AttributeReID2,AttributeReID3,AttributeReID4,NewAtt1,NewAtt2}, \KK{only} a few work \cite{AttributeSearch1} has discussed attribute-image Re-ID problem.
The work in \cite{AttributeSearch1} is to form attribute-attribute matching. However, 
despite the improvement on performance achieved by attribute prediction methods \cite{DeepMAR}, directly retrieving people according to their predicted attributes is still challenging, because the attribute prediction methods are still not robust to the cross-view condition changes like different lighting conditions and viewpoints.

\looseness=-1
In this work, for the first time, we \K{present} extensive investigation on the attribute-image Re-ID problem under an adversarial framework. Rather than directly predicting attributes of an image, we cast the cross-view attribute-image matching as cross-modality matching by an adversarial learning problem.

\subsection{\yin{Cross-modality Retrieval}}
\looseness=-1
Our work is related to cross-modality content search, which aims to bridge \K{the gap} between different modalities \cite{CCA,RCCA,DeepCCA,rankingLoss1}. \CameraReady{The most traditional and practical solution to this task is  Canonical Correlation Analysis (CCA) \cite{CCA,RCCA,DeepCCA}, which projects two modalities into a common space that maximizes their correlation. }Ngiam et al.  and Feng et al. also applied autoencoder-based methods to model the cross-modality correlation \cite{autoencoder1,autoencoder2}, and Wei et al. proposed a deep semantic matching method to address the cross-modality retrieval problem with respect to samples annotated with one or multiple labels\cite{DeepSM}. Recently, A. Eisenschtat and L. Wolf have designed a novel model of two tied pipelines that maximize the projection correlation using an Euclidean loss, which achieves state of the art results in some datasets \cite{2WayNet}.
\CheckYinzhou{{Two most related works to ours are proposed in \cite{Li01,CMCE}, which retrieve pedestrian images using language descriptions.}}
% A most related work to ours was proposed by\cite{Li01}, which applies a neural attention mechanism to retrieve pedestrian images using language descriptions.
Compared with this setting, our attribute-image Re-ID has its own strength in embedding more pre-defined attribute descriptions for obtaining better performance.

 \subsection{Distribution Alignment Methods}

\looseness=-1
\CheckYinzhou{The adversarial model employed in this work is in line with the GAN methods\cite{DomainGAN2,GAN-CLS,OriGAN}, which has its strength in \yin{distribution} alignment by a two player \yin{min-max} game.}
\K{As different modalities follow different distributions, our cross-modality attribute-image Re-ID problem is also related to the distribution alignment problem.}
% The adversarial model employed in this work is in line with the GAN methods\cite{DomainGAN2,GAN-CLS} has its strength in domain alignment by a two player minimax game. 
For performing \yin{distribution} alignment, there are other techniques called domain adaptation techniques \cite{MMD1,MMD2,DA1}. In domain adaptation, to align the distribution of data from two different domains, several works \cite{MMD1,MMD2} apply MMD-based loss, which minimize the norm of difference between means of the two distributions. Different from these methods, the deep Correlation Alignment(CORAL) \cite{DeepCorr} method proposed to match both the mean and covariance of the two distributions. 
%In this work, we examine all these approaches and find the adversarial modelling is more suitable for the attribute-image Re-ID, which could be due to its more powerful ability on aligning more discrepant data distributions.
\K{Our work is different from these methods as our framework not only bridges the gap between the two largely discrepant modalities, but also keeps the semantic consistency across them}.

%Our work is different from these methods as to address the attribute based ReID, our model not only aligns distribution structure to transfer knowledge across modalities, but also keeps consistency in the semantic level to correlate different modalities better.

\begin{figure}
\begin{center}
\includegraphics[width=1.0\linewidth,height=0.6\linewidth]{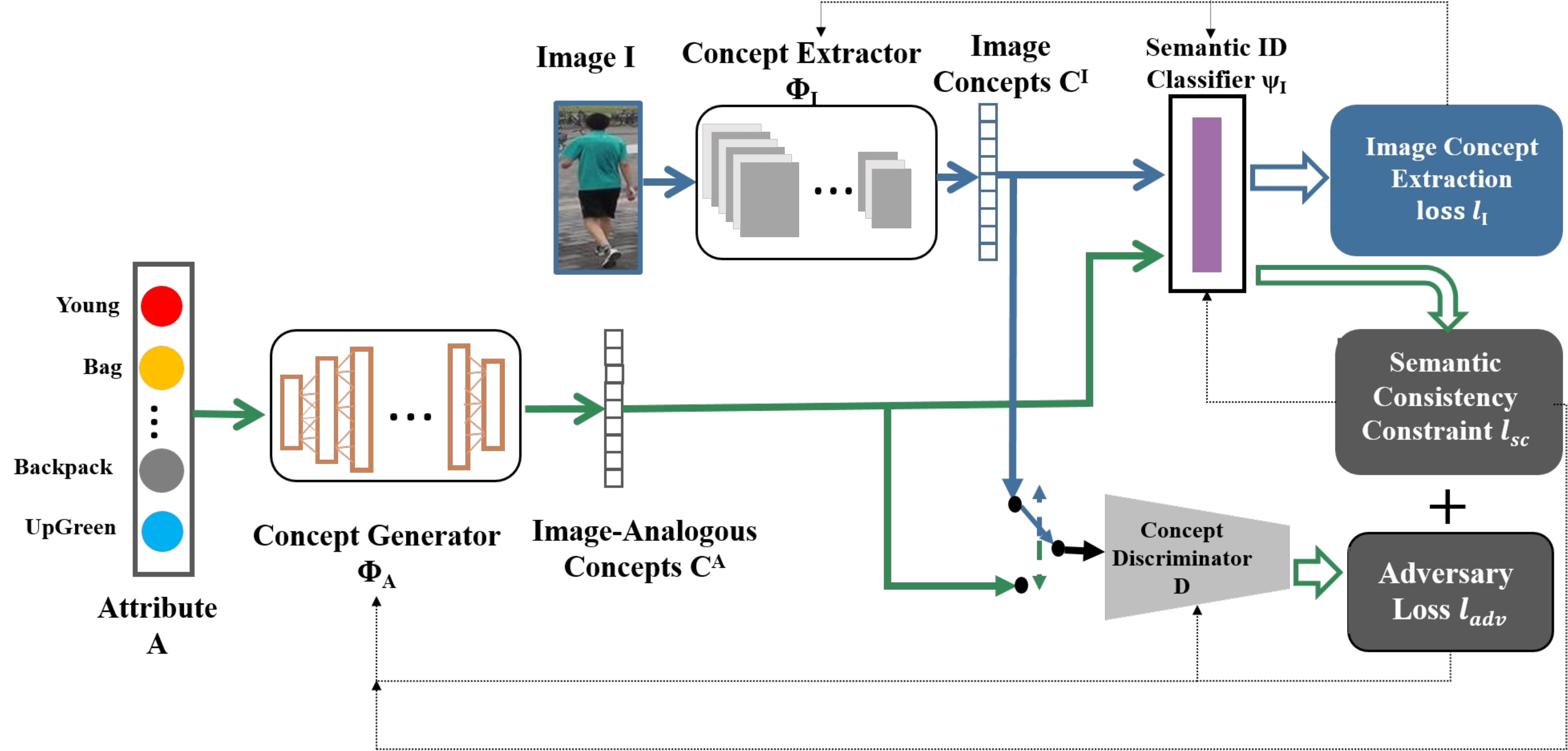}
\end{center}
\vspace{-0.5cm}
   \caption{The structure of our network. \CameraReady{It consists of two branches: the image branch (top blue branch) learns to extract semantic concepts from images, and the attribute branch (bottom green branch) learns to generate image analogous concepts from attributes. Dash lines represent the gradient flow of the three objectives that we propose.}
   %It aims to generate image-analogous concept for an attribute input optimized by three \yin{loss} functions: image concept extraction loss, semantic consistency constraint, and adversary loss.
 See Sec. \ref{section:architecture} for details about the network architecture.}
%   consists of two pipelines. The top line of this framework learns attribute embedding $\Phi_A$, and the bottom line learns image embedding $\Phi_I$. Classifiers with shared weights that are used to classify semantic IDs are applied in both pipeline to maintain the semantic consistency across modalities. The attribute pipeline acts as a generator in the generative adversary architecture, and tries to fool the concept discriminator in the middle, which is trained to estimate the probability of whether its input is from the image pipeline.}
\label{network}
\end{figure}

%------------------------------------------------------------------------
\section{Attribute-image Person Re-ID}
\looseness=-1
Given an attribute description $A_i$, attribute-image Re-ID aims at re-identifying the matched pedestrian images $I_i$ from an image database $\mathcal{I} = \{I_i\}_{i=1}^{N}$ captured under real surveillance environment, where $N$ is the size of $\mathcal{I}$. Since different person images could have the same attribute description, the attribute-image Re-ID uses Semantic ID (i.e., attribute description) to group person images. \ancong{That is, people with the same attribute description are of the same semantic ID.}
%, since all persons matching the query attribute description are the targets, regardless of person identities.}

\looseness=-1
The goal of our method is to learn two mappings $\Phi_I$ and $\Phi_A$ that respectively map person images and high-level semantic attributes into a joint discriminative space, which could be regarded as the concept space as mentioned. That is, $C^I_i = \Phi_I (I_i)$ and $C^A_i = \Phi_A (A_i)$, where 
$C^I_i$ and $C^A_i$ are the mid-level concept that is generated from the image $I_i$ and attribute $A_i$, respectively.
To achieve this, we form an image embedding network by CNN and an attribute-embedding network by a deep fully connected network. We parameterize our model by $\mathbf{\Theta}$, and obtain $\mathbf{\Theta}$ by optimizing a concept generation objective $L_{concept}$. Given training pairs of images and attributes $(I_i, A_i)$, the optimization problem is formulated as
\begin{equation}
\begin{aligned}
\mathop{\min}_{\mathbf{\Theta}} L_{concept} &= \frac{1}{N}\sum\limits_{i=1}^{N}l_{concept}  (\Phi_I (I_i),\Phi_A (A_i)).
%&= \frac{1}{N}\sum\limits_{i=1}^{N}l_{concept}  (C^I_i,C^A_i).
\end{aligned}
\end{equation}

In this paper, we design $l_{concept}$ as a combination of several loss terms, each of which formulates a specific aspect of consideration to jointly formulate our problem.
The first consideration is that the concepts we extract from the low-level noisy person images should be semantically discriminative.
We formulate it by image concept extraction loss $l_I$.
The second consideration is that image-analogous concepts $C^A$ generated from attributes and image concepts $C^I$ should be homogeneous. Inspired by the powerful ability of generative adversary networks to close the gap between heterogeneous distributions, we propose to embed an adversarial learning strategy into our model.
This is modelled by a concept generating objective \CheckYinzhou{$l_{CG}$}, which aims to generate concepts not only discriminative but also homogeneous with concepts extracted from images.
Therefore, we have
\begin{equation}
l_{concept}= l_I + l_{CG} \label{4}.
\end{equation}
In the following, we describe each of them in detail.

\subsection{Image Concept Extraction}\label{semanticID}
\looseness=-1
Our image concept extraction loss $l_I$ is based on softmax classification on the image concepts $\Phi_I(I)$.
Since our objective is to learn semantically discriminative concepts that could distinguish \KK{different attributes rather than specific persons},
we re-assign semantic IDs $y_i$ for any person image $I_i$ according to its attributes rather than real person IDs, which means different people with the same attributes have the same semantic ID.
We define the image concept extraction loss as a softmax classification objective on semantic IDs. Denoting \ancong{$\Psi_{I}$} as the classifier and $I$ as the input image, the image concept extraction loss is the negative log likelihood of predicted scores \ancong{$\Psi_I(\Phi_I(I))$}:

\yinzhou{
\begin{eqnarray}\label{eq:image_cls}
l_{I} = \sum_{i} -\log \Psi_I(\Phi_I(I_i))_{y_{I_i}},
\end{eqnarray}
}
%where \yinzhou{\textbf{$\mathbb{I}$ denotes the one hot label vector,}} $i$ refers to the ground truth semantic ID of $I$.
where $I_i$ is the $i^{th}$ input image, $y_{I_i}$ is the semantic ID of $I_i$ and $\Psi_I(\Phi_I(I_i))_{k}$ is the $k^{th}$ element of $\Psi_I(\Phi_I(I_i))$.
% \begin{eqnarray}
% l_I (C^I_i) = -\log  (\frac{exp  (\mathbf{f}^I_{i, y_i})}{\sum_j exp  (\mathbf{f}^I_{i, j})})
% \label{equation:imageConcept}
% \end{eqnarray}
% where the $j$-th entry $\mathbf{f}^I_{i, j}$ is the unnormalized log probability of $I_i$ having the semantic ID $j$.

\subsection{Semantic-preserving Image-analogous Concept Generation}
\vspace{0.1cm}
\noindent{\textbf{Image-analogous Concept Generation.}}
We regard $\Phi_A$ as a generative process, just like the process of people generating an imagination from an attribute description.
As the semantically discriminative latent concepts could be extracted from images, they can also provide information to learn the image-analogous concepts $\Phi_A(A)$ for attributes as a guideline.

\looseness=-1
\K{Mathematically}, the generated image-analogous concepts should \K{follow} the same distribution as image concepts, i.e., $P_I(C)=P_A(C)$, where $C$ denotes a concept in the joint concept space of $\Phi_I(I)$ and $\Phi_A(A)$ and $P_I$, $P_A$ denote the distribution of image concepts and image-analogous concepts, respectively. We \K{learn} a function $\hat{P_I}$ to approximate image concept distribution $P_I$, and force the image-analogous concepts $\Phi_A(A)$ \K{to follow} distribtion $\hat{P_I}$. It can be achieved by an adversarial training process of GAN, in which discriminator $D$ is regarded as $\hat{P_I}$ and the generator $G$ is regarded as image-analogous concept generator $\Phi_A$.

In the adversary training process, we design a network structure (see Sec. \ref{section:architecture}) and train our concept generator $\Phi_A$ with a goal of fooling a skillful concept discriminator $D$ that is trained to distinguish the image-analogous concept from the image concept,
so that the generated image-analogous concept is aligned with the image concept.
We design the discriminator network $D$ with parameters $\theta_D$ and denote the parameters of $\Phi_A$ as $\theta_G$.
The adversarial min-max problem is formulated as
%by alternately optimizing $D$ and $\Phi_A$:
\begin{equation}
\begin{split}
\min_{\theta_G}\max_{\theta_D} V(D,G) = &\mathbb{E}_{I\sim p_I}[\log D(\Phi_I(I))]+ \\
                 &\mathbb{E}_{A\sim p_A}[\log (1-D(\Phi_A(A)))]\label{1}.
\end{split}
\end{equation}
\looseness=-1
The above optimization problem is solved by iteratively optimizing $\theta_D$ and $\theta_G$.
Therefore, the objective can be decomposed into two loss terms $l_{adv}^G$ and $l_{adv}^D$, which
are for training the concept generator $\Phi_A$ and the discriminator $D$, respectively. Then the whole objective during adversary training $l_{adv}$ could be formed by:
% \begin{align}
% &l_{adv} =  \lambda_D l^{D}_{adv} + \lambda_G l^{G}_{adv},\\
% &\text{where} 
% \ \CheckYinzhou{l_{adv}^G = -\mathbb{E}_{A \sim p_A}[\log D(\Phi_A(A))]},\nonumber\\
% &\ \ \ \ \  \ \ \ \ \ \  
%  \CheckYinzhou{l_{adv}^D = -\mathbb{E}_{I \sim p_I}[\log D(\Phi_I(I))]}\nonumber \\
%  &\ \ \ \ \  \ \ \ \ \ \ \ \ \ \ \ \ \ \ \ \ \  \ \ \ \ \ \  \CheckYinzhou{-\mathbb{E}_{A \sim p_A}[\log (1-D(\Phi_A(A)))]}\nonumber.
% \end{align}
\K{
\begin{align}
l_{adv} =  \lambda_D l^{D}_{adv} + \lambda_G l^{G}_{adv},
\end{align}
where 
\begin{align}
l_{adv}^G = &-\mathbb{E}_{A \sim p_A}[\log D(\Phi_A(A))],\nonumber\\  
l_{adv}^D = &-\mathbb{E}_{I \sim p_I}[\log D(\Phi_I(I))] \nonumber \\
&-\mathbb{E}_{A \sim p_A}[\log (1-D(\Phi_A(A)))]. \nonumber
\end{align}
}
\vspace{0.1cm}

% \subsection{Semantic Consistency Constraint}
\noindent{\textbf{Semantic Consistency Constraint.}
\looseness=-1
The adversarial learning pattern $l_{adv}$ is important for generator $\Phi_A$ to generate image-analogous concept with the same distribution of image concept $\Phi_I(I)$. Furthermore, we should generate meaningful concepts preserving the semantic discriminability of the attribute modality, {i.e., $P^{sid}_I(C)=P^{sid}_A(C)$}, where $P_I^{sid}$ and $P_A^{sid}$ denote the distributions of image concepts and image-analogous concepts of semantic ID $sid$. {If we analyze the image concept extraction loss $l_I$ in Equation (\ref{eq:image_cls}) independently, $\Psi_I$ can be regarded as a function to approximate a set of distributions $P_I^{sid}(C)$ for each semantic ID $sid$. With the assumption that the generated image-analogous concepts should be in the same concept space as image concepts, $\Psi_I$ is shared by image concept extraction and image-analogous concept generation, so as to guarantee identical distribution of two \yin{modalities} in semantic ID level.}
We integrate a semantic consistency constraint $l_{sc}$ using the same classifier for image concept {$\Psi_I$}:
\begin{eqnarray}
%{l_{sc} = -\log\{(\sum_j \exp(\Psi_I(\Phi_A(A)))_j)^{-1}\exp(\Psi_I(\Phi_A(A)))_i\}.}
l_{sc} = \sum_{i} -\log \Psi_I(\Phi_A(A_i))_{y_{A_i}},
\end{eqnarray}
where $A_i$ is the $i^{th}$ input attribute, $y_{A_i}$ is the semantic ID of $A_i$ and $\Psi_I(\Phi_A(A_i))_{k}$ is the $k^{th}$ element of $\Psi_I(\Phi_A(A_i))$.
Thus the overall concept generating objective for attributes \CheckYinzhou{$l_{CG}$} becomes the sum of $l_{adv}$ and $l_{sc}$:
\begin{eqnarray}
\CheckYinzhou{l_{CG}} = l_{adv}+l_{sc}.
\end{eqnarray}
By this way, we encourage our \KK{generative model to generate a more homogeneous image-analogous concept space, while at the same time correlate image-analogous concepts with semantically matched image concepts by maintaining semantic discriminability in the learned space.}

\subsection{The Network Architecture}\label{section:architecture}
\looseness=-1
Our network architecture is shown in Figure \ref{network}. \ja{Firstly,} \K{the concept generator \ja{is particularly designed to have} multiple fully connected layers in order to obtain enough capacity to generate image-analogous concepts which are highly heterogeneous from the input attribute. Details are shown in Table \ref{attributeNet}.}
 \CameraReady{Secondly, our concept discriminator is also a combination of fully connected layers, each followed by batch normalization and leaky reLU, except for the output layer, which is processed by the Sigmoid non-linearity.}
 \ja{Finally, the} \K{concept extractor} is obtained by removing the last Softmax classification layer of Resnet-50 and adding a \K{128-D fully connected layer. We regard the feature produced by the FC layer as the image concept. Note that the dimension of the last layer in the concept generator is also set to 128.} 

\looseness=-1
As introduced above, we impose the semantic consistency constraint on attribute and thus we pass image-analogous concepts into \K{the same Semantic ID classifier as that for image concepts}.
%the the concept extraction loss of image and the concept generation loss of attributes share the same high level classification process, and thus the two pipelines of our model concatenate a shared classification layer.
At the inference stage, we rank the gallery pedestrian image concepts \yin{$C^I$} according to their cosine distances to the query image-analogous concepts \yin{$C^A$} in the latent embedding space.

\begin{table}
\begin{center}
\scriptsize
\begin{tabular}{|l|c|}
\hline
Structure & Size \\
\hline\hline
fc1 & $attributeSize \times 128$\\
BatchNormalization & 128 \\
ReLU & 128\\
fc2 & $128\times 256$\\
BatchNormalization & 256 \\
ReLU & 256\\
fc3 & $256 \times 512$\\
BatchNormalization & 512 \\
ReLU & 512\\
fc4 & $512 \times embeddingSize$\\
Tanh & \CheckYinzhou{$embeddingSize$}\\
\hline
\end{tabular}
\end{center}
\caption{The structure of our network' attribute part. Fc means fully connected layers. 
%Different dataset has different attribute size and semantic ID size. 
128 is set to be the embedding size in our work.}
\label{attributeNet}
\end{table}

\vspace{0.1cm}

\noindent \textbf{Implementation Details}.
%Our source code is modified from Open-ReID\footnote{https://cysu.github.io/open-reid/}, using the Pytorch framework. 
\looseness=-1
We first pre-trained our image network for 100 epochs using the semantic ID, with an adam optimizer \cite{Adam} with learning rate 0.01, momentum 0.9 and weight decay 5e-4. After that, we jointly train the whole network. We set $\lambda_G$ in Eq.  (\ref{4}) as 0.001, and $\lambda_D$ as 0.5, which will be discussed in Section \ref{section:evaluations}. The total epoch was set to 300. During training, we set the learning rate of the attribute branch to 0.01, and set the learning rate of the image branch to 0.001 because it had been pre-trained. \CameraReady{The batch size of training is 128 and the setting of optimizer is the same as that of pre-training}. \KK{Hyper-parameters} are fixed in comparisons across all the datasets.

%By iteratively optimizing $l_{adv}^D$ and $l_{adv}^G$ in Eq. (\ref{equation:ladv}), just like what was done in classical GAN works, we encourage our generation model to generate homogeneous image-analogous concept structure.
%And meanwhile, our model could correlate image-analogous concepts with semantically matched image concepts by maintaining semantic discriminability in the learned joint space.

%-------------------------------------------------------------------------
%-------------------------------------------------------------------------

%-------------------------------------------------------------------------
\section{Experiments}
%To evaluate the effectiveness of our approach, we conduct several groups of experiments on three large scale pedestrian attribute datasets. %In this section, we first introduce statistics of the used datasets, and detail the implementations in section 4.1 and 4.2. Then we compare our joint alignment framework/**YinZhou:Or other names?**/ with the widely used CCA methods and the intuitive attribute prediction solutions in section 4.3. At last, Section 4.4 illustrates the effectiveness of our proposed GAN-based structure alignment.

\subsection{Datasets and Settings}
\looseness=-1
\noindent\textbf{Datasets.}
We evaluate our approach and compare with related methods on three benchmark datasets, including Duke Attribute \cite{lin2017improving}, Market Attribute  \cite{lin2017improving}, and PETA \cite{PETA}. 
We tried to follow the setting in literatures.
The Duke Attribute dataset contains 16522 images for training, and 19889 images for testing. Each person has 23 attributes. We labelled the images using semantic IDs according to their attributes. As a result, we have 300 semantic IDs for training and 387 semantic IDs for testing. Similar to Duke Attribute, the Market Attribute also has 27 attributes to describe a person, with 12141 images and 508 semantic IDs in the training set, and 15631 images and 484 semantic IDs in the test set. For PETA dataset, each person has 65 attributes  (61 binary and 4 multi-valued). 
%While PETA is not given for Re-ID before, 
We used 10500 images with 1500 semantic IDs for training, and 1500 images with 200 semantic IDs for testing.
%, while we could still achieve satisfactory results.

\vspace{0.1cm}

\noindent\textbf{Evaluation Metrics.} We computed both Cumulative Match Characteristic (CMC) and mean average precision (mAP) as metrics to measure performances of the compared models.

\begin{table*}
\setlength{\belowcaptionskip}{-0.2cm}
\scriptsize
\begin{center}
\begin{tabular}{|l|c|c|c|c|c|c|c|c|c|c|c|c|}
\hline
\multirow{2}{*}{Method}
&\multicolumn{4}{|c|}{Market}&\multicolumn{4}{|c|}{Duke}&\multicolumn{4}{|c|}{PETA}  \\
\cline{2-13}
&rank1&rank5&rank10&mAP&rank1&rank5&rank10&mAP&rank1&rank5&rank10&mAP \\
\hline\hline
DeepCCAE \cite{DCCAE}&8.12&23.97&34.55&9.72&33.28&59.35&67.64&14.95&14.24&22.09&29.94&14.45\\
\hline
DeepCCA \cite{DeepCCA}&29.94&{\color{blue}\textbf{50.70}}&{\color{blue}\textbf{58.14}}&17.47&36.71&58.79&65.11&13.53&14.44&20.77&26.31&11.49\\
\hline
2WayNet \cite{2WayNet}& 11.29&24.38&31.47&7.76&25.24&39.88&45.92&10.19&23.73&38.53&41.93&15.38 \\
\hline
DeepMAR \cite{DeepMAR}&13.15&24.87&32.90&8.86&36.60&57.70&67.00&14.34&17.80&25.59&31.06&12.67\\
\hline
CMCE \cite{CMCE}&35.04&{\color{red}\textbf{50.99}}&56.47&{\color{red}\textbf{22.80}}&39.75&56.39&62.79&15.40&31.72&39.18&48.35&26.23\\
\hline\hline
ours w/o adv&33.83&48.17&53.48&17.82&39.30&55.88&62.50&15.17&36.34&48.48&53.03&25.35\\
\hline
ours w/o
sc&2.08&4.80&4.80&1.00&5.26&9.37&10.87&1.56&3.43&4.15&4.15&5.80\\
\hline
ours w/o adv+MMD&34.15&47.96&57.20&18.90&41.77&{\color{red}\textbf{62.32}}&{\color{blue}\textbf{68.61}}&14.23&{\color{red}\textbf{39.31}}&48.28&{\color{blue}\textbf{54.88}}&{\color{red}\textbf{31.54}}\\
\hline
ours w/o adv+DeepCoral &{\color{blue}\textbf{36.56}}&47.61&55.92&20.08&{\color{blue}\textbf{46.09}}&{\color{blue}\textbf{61.02}}&68.15&{\color{red}\textbf{17.10}}&35.62&{\color{blue}\textbf{48.65}}&53.75&27.58\\
\hline\hline
ours&{\color{red}\textbf{40.26}}&49.21&{\color{red}\textbf{58.61}}&{\color{blue}\textbf{20.67}}&{\color{red}\textbf{46.60}}&59.64&{\color{red}\textbf{69.07}}&{\color{blue}\textbf{15.67}}&{\color{blue}\textbf{39.00}}&{\color{red}\textbf{53.62}}&{\color{red}\textbf{62.20}}&{\color{blue}\textbf{27.86}}\\
\hline
\end{tabular}
\end{center}
\caption{Comparison results on the three benchmark datasets. Performances are measured by the rank1, rank5 and rank10 matching accuracy of the cumulative matching curve, as well as mAP.
The best performances are indicated in {\color{red}\textbf{red}} and the second indicated in {\color{blue}\textbf{blue}}.}
\label{ComparingTable}
\end{table*}

\vspace{0.1cm}

\subsection{Evaluation on the Proposed Model}\label{section:evaluations}
\looseness=-1
\noindent \textbf{Adversarial vs. Other \yin{\ja{Distribution Alignment}} Techniques.} For our \K{attribute-image Re-ID}, we employ the adversarial technique to make the image-analogous concepts generated from attribute aligned with the image concepts. While CCA is also an \KK{option and will be discussed when comparing our method with DCCA} later, we examine whether other widely used alignment methods can work for our problem. We consider the MMD objective, which minimize difference between means of two distributions, and DeepCoral \cite{DeepCorr}, which matches both mean and covariance of two distributions, as traditional and effective distribution alignment baselines. Since their original models cannot be directly applied, we modify our model for comparison, that is we compare with 1) our model without the adversary learning but with an MMD objective(ours w/o adv+MMD); 2) our model without the adversary learning but with Coral objective(ours w/o adv+DeepCoral). We also provide the baseline that adversarial learning is not presented, denoted as ``ours w/o adv''.

Compared with the model that does not use adversarial learning (ours w/o adv), all the other baselines including our adversary method perform clearly better.
Among all, the adversary learning framework generally performs better (with the best and second best performance)
% , although the performance is sometimes slightly inferior on Duke as compared with (ours w/o adv+DeepCoral) 
as shown in Table \ref{ComparingTable}.

\vspace{0.1cm}

\looseness=-1
\noindent \textbf{With vs. Without Semantic Consistency Constraint}. In our framework, we tested our performance when the semantic consistency constraint is not used, denoted as ``ours w/o sc''. As reported in Table \ref{ComparingTable}, without semantic consistency constraint the performance drops sharply. This is because although the \yin{distributions of two modalities} are \ja{aligned}, the corresponding pair is not correctly matched. Hence, the semantic consistency constraint actually regularizes the adversarial learning to avert this problem. As shown, with semantic consistency constraint but without adversarial learning (i.e., ``ours w/o adv'') \yin{our model} clearly performed worse than our full model.
% and is even not inferior to other alignment technique 
All \K{the observations suggest} the generic adversarial model itself does not directly fit the task of aligning two \yin{modalities} which are highly discrepant, but the regularized one by semantic consistency constraint does.

\vspace{0.1cm}

\looseness=-1
\noindent \textbf{A2Img vs. Img2A}. 
In our framework, we currently use the adversarial loss to align the generated image-analogous concept of attribute towards image concept, we call such case generation from attributes to image (A2Img). We now provide comparative results on generation from image to attriburtes (Img2A). As reported in Table \ref{AblationTable},
we find that Img2A is also effective, which even outperforms A2Img on the PETA dataset. But on larger datasets Market and Duke, A2Img performs better. The reason may be that the distribution of semantic IDs is much sparser than the distribution of images. Thus, estimating the manifold of images from the training data is more reliable than estimating that of attributes. But in PETA, the number of images is relatively small while semantic IDs are relatively abundant compared with the other two datasets. Moreover, PETA also has more complicated sceneries and larger number of attribute descriptions, which are more challenging for images to learn discriminative concepts. Thus learning generated attribute-analogous concepts and aligning with attribute concepts provides more discriminative information, and Img2A performs better on PETA.

\begin{table}
\setlength{\belowcaptionskip}{-0.2cm}
\begin{center}
\scriptsize
\begin{tabular}{|c|c|c|c|}
\hline
Method&Market&Duke&PETA  \\
\hline\hline
\ja{Ours (i.e. A2Img)}&\textbf{40.3}&\textbf{46.6}&39.0\\
\hline
Img2A (reverse of the proposed)&36.0 &43.7&\textbf{43.6}\\
% \hline
% Both &38.6&41.3&34.4\\
\hline
Real Images&8.13&20.01&19.85\\
%\hline
%w/o adv &33.8 &39.3&36.5 \\
\hline
\end{tabular}
\end{center}
\caption{The rank1 matching accuracy of some variants of our model.
``A2Img'' denotes our model which generates concepts from attributes.
``Img2A'' does the reverse of ``A2Img''.
%the model in which we simply exchange the position of $C^I_i$ and $C^A_i$.
``Real Images'' denotes the model which generates images  (rather than concepts) for attributes.
%``w/o adv'' denotes the model in which the adversarial loss is removed.
}
\label{AblationTable}
\end{table}

\vspace{0.1cm}

\looseness=-1
\noindent \textbf{Generation in Concept Space or in Image Space}.
What if our model generates images instead of concepts, according to the attributes? We study how the generated image-analogous pattern (whether concepts or images) affects the effectiveness of our model. \CameraReady{To this end, we use the conditional GAN in \cite{GAN-CLS} to generate fake image, which have aligned structure with real images, from our semantic attributes and a random noise input. We have modified some input dimension and added some convolution and deconvolution layers in \cite{GAN-CLS} to fit our setting. Firstly we train the generative models for 200 epochs, and then the classification loss is
added for another training of 200 epochs.}

%Specifically, the code we use to generate fake images is from Reed et.al \cite{GAN-CLS}, where we have modified some input dimension and added some convolution and deconvolution layers to fit our setting. We train the generative models for 200 epochs, and then the classification loss is added for another training of 200 epochs, where the trade-off parameter of the generative loss $\lambda_G$ is also set to 0.001 as our settings.
\looseness=-1
We find the retrieval performance is worse than our original model, as shown in Table \ref{AblationTable}.
\K{
This is probably because generating the whole pedestrian image introduces some noise, which is harmful in discriminative tasks like attribute-image Re-ID. In contrast, generating concepts which are relatively ``clean'' can avoid introducing unneccesary noise. Thus, generating image-analogous concepts in the discriminative concept space is more effective.}

\subsection{Comparison with Related Work}
\looseness=-1
\noindent \textbf{Comparing with Attribute Prediction Method.} 
As mentioned above, an intuitive method of attribute-image Re-ID is to predict attributes from person images and perform the matching between predicted attributes and query attributes. We compare our model with a classical attribute recognition model DeepMAR \cite{DeepMAR}, which formulates attribute recognition as a muti-task learning problem and acts as an off-the-shelf attribute predictor in our experiment.
As shown in Table \ref{ComparingTable}, our model outperforms DeepMAR, and it is because DeepMAR still suffers from the \K{problem of indiscriminative predicted attributes.} %resulted by the model capacity and data noise. 
Different from DeepMAR, we choose to learn latent representations as the bridge between the two modalities, where we successfully avert the problem caused by attribute prediction and learn more discriminative concepts using adversary training.

\vspace{0.1cm}

\noindent \textbf{Comparing with Cross Modality Retrieval Models.} Since our problem is essentially a cross-modality retrieval problem, we compare our model with the typical and commonly used Deep canonical correlation analysis (DCCA)  \cite{DeepCCA}, Deep canonically correlated autoencoders (DCCAE) \cite{DCCAE} and a state-of-the-art model 2WayNet \cite{2WayNet}.
Deep CCA applies the CCA objective in deep neural networks in order to maximize the correlation between two different modalities. DCCAE\cite{DCCAE} jointly models the cross-modality correlation and reconstruction information in the joint space learning process. 2WayNet is a \CameraReady{recently proposed} \K{two-pipeline model which maximizes sample correlations}.

\looseness=-1
\K{We show the comparative results in Table \ref{ComparingTable}. \CameraReady{From Table \ref{ComparingTable}, }
we can observe that our model outperforms all the cross modality retrieval baselines on all three datasets by large margins. This is partially because our model not only learns to close the gap between the two modalities in the joint concept space, but also keeps the semantic consistency of the extracted and generated concepts.}

\looseness=-1
In addition, we compare our model with the most related one, i.e., the cross modality cross entropy (CMCE) model \cite{CMCE}, which achieved a state-of-the-art result in text-based person retrieval.
\K{
We train the CMCE model with semantic ID for fair comparison. The results in Table \ref{ComparingTable} show that our model is comparable (\ja{on} Market) or more effective (\ja{on} Duke and PETA) for the attribute-image Re-ID problem.}
}

\subsection{Further Evaluations}

%First, we further evaluate the influence of using semantic identities (denoted as ours(Real ID)).
%rather %than truly pedestrian identities (ours w/o adv vs. ours w/o adv+Real ID). 
%As the 109 attribute items in PETA is enough to describe a unique person, the semantic identities in that dataset are almost the same as Real identities, we only conducted the experiment on the other two datasets. From Table \ref{ComparingTable} we find that although we could also learn semantically discriminative concepts using real identities, using semantic identities performs better.% in all metrics.

\K{Finally we present some further evaluations of our model. We first evaluate the effects of two important hyper-parameters $\lambda_D$ and $\lambda_G$. We present the results on the Duke Attribute dataset in Figure \ref{lambdaCurve}. The trends are similar on other datasets, and therefore Figure \ref{lambdaCurve} might be useful for determining the hyper-parameters \CameraReady{on other datasets}.}

\begin{figure}
\begin{center}
\includegraphics[width=1\linewidth]{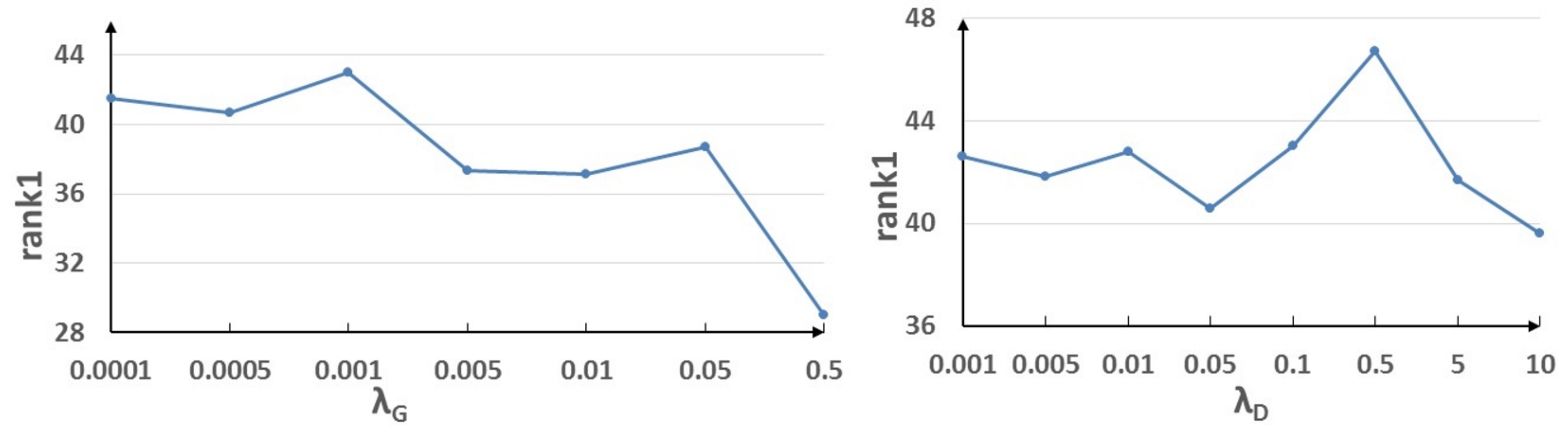}
\end{center}
   \caption{Results of experiment on the trade-off parameters $\lambda_G$ and $\lambda_D$. We firstly set $\lambda_D$ to 1 and change the value of $\lambda_G$, and get the results in the left image. Then we chose our best $\lambda_G$=0.001 in our experiments and change $\lambda_D$ on the right.}
\label{lambdaCurve}
\end{figure}

\begin{figure}[t]
\setlength{\belowcaptionskip}{-0.2cm}
\begin{center}
\includegraphics[width=1\linewidth,height=0.36\linewidth]{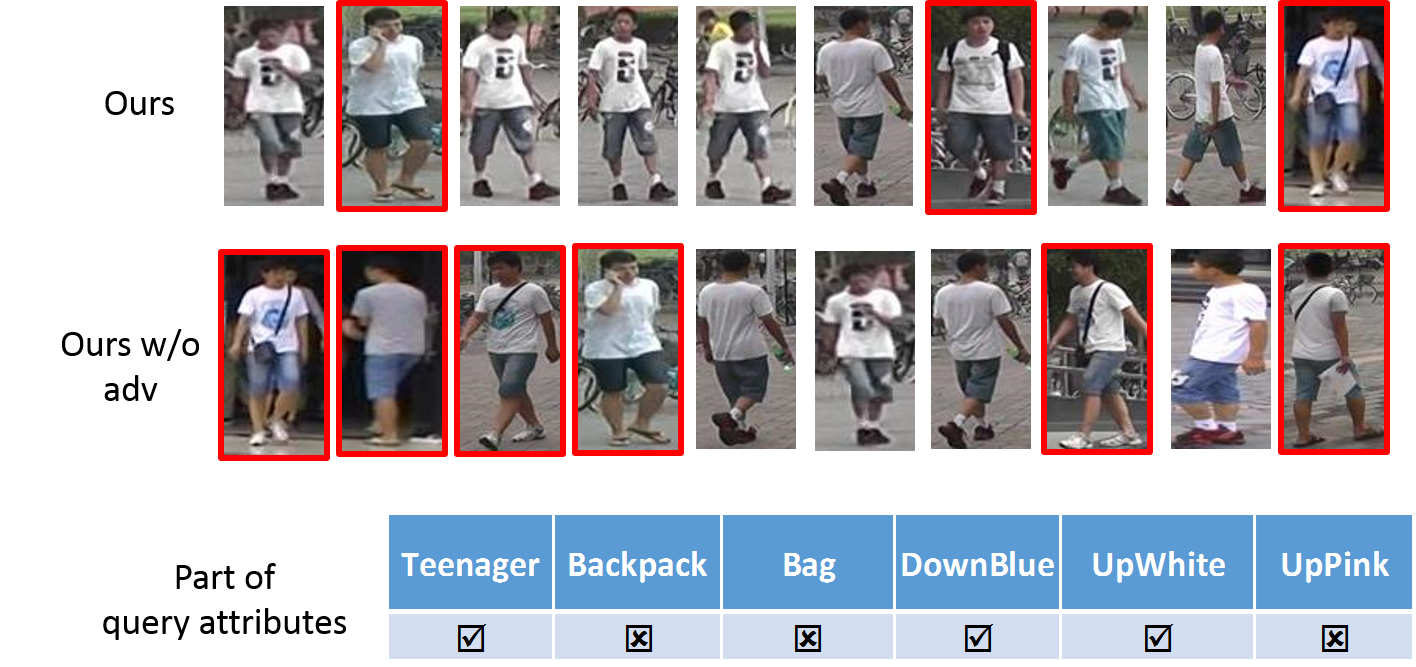}
\end{center}
   \caption{Qualitative example in Market Attribute Dataset. The first row shows the results of our proposed method and the second are about a baseline. To save space, we only list 6 attribute items among all the 27 ones in Market Attribute in the third row. The inaccurately retrieved samples are marked by red rectangles in the figure.}
\label{qualitative}
\end{figure}

%-------------------------------------------------------------------------

\looseness=-1
Secondly, we conduct qualitative evaluations on our proposed model. Figure \ref{qualitative} shows examples of top-10 ranked images according to a query attribute description \K{from} the Market Attribute dataset. 
%As there are 27 attribute items in Market Attribute, for space saving, we only list 6 attribute items which are helpful for readers to distinguish wrongly retrieved samples from the right retrieved ones. 
We find that fine-grained features of pedestrian images (e.g. stride of a backpack) are the main reasons that cause mistakes in our baseline (see ours w/o adv in the second row of Figure \ref{qualitative}). But with the adversarial objective, our model could get an intuition and generate the concept of what a person wearing a backpack would look like, and then concentrate more on possible fine-grained features.

\section{Conclusion}
\looseness=-1
\ja{The attribute-image Re-ID problem is a cross-modal matching problem that is realistic in practice, and it differs notably from the previous attribute-based person Re-ID problem that is still essentially an image-image Re-ID problem.} In this work, we have identified its challenge through the experiments on three datasets. We have shown that an adversarial framework regularized by a semantic consistency constraint is so far the most effective way to solve the attribute-image Re-ID problem. Also, by our learning, we find that under the regularized adversarial learning framework, it is more useful to learn image-analogous concept from inquired attributes and make it aligned with the corresponding real image's concept, as compared \yin{with} its reverse. 

\section*{Acknowledgements}
This work was supported partially by the National Key Research and Development Program of China (2016YFB1001002),NSFC(61522115, 61661130157, 61472456, U1611461, 61573386), Guangdong Province Science and Technology Innovation Leading Talents (2016TX03X157),  the Royal Society Newton Advanced Fellowship (NA150459), Guangdong NSF (2016A030313292), Guangdong Program (2016B030305007, 2017B010110011).
%, Guangzhou Science and Technology Project (201804010435).

\bibliographystyle{named}
\bibliography{egbib,reid_bib_conference,relatedWork}

\end{document}